\documentclass[letterpaper, 10pt, conference]{ieeeconf}  

\IEEEoverridecommandlockouts                              

\usepackage{multirow}
\usepackage{amsmath, amsfonts}
\usepackage{xcolor}
\usepackage{subcaption}
\usepackage[utf8]{inputenc}
\usepackage{tikz, tabularx}
\usepackage{overpic}
\usepackage{lipsum}
\usepackage{placeins}
\usepackage{graphicx}
\usepackage{booktabs}
\usepackage{censor}
\usepackage{caption}

\title{\LARGE \bf
Proactive Robot Assistance via Event-Driven Reasoning over State Changes
}
\author{Fengkai Liu$^{1}$, Hao Su$^{1}$, Haozhuang Chi$^{2}$, Rui Geng$^{1}$, Congzhi Ren$^{1}$, Xuqing Liu$^{1}$, Chenfei Xu$^{1}$, \\ Yuichi Ohsita$^{1}$ and Liyun Zhang$^{3 \dagger }$
\\[0.4em]
{\small $^{1}$The University of Osaka \quad $^{2}$Nanyang Technological University} \small \quad $^{3}$The University of Tokyo\\[0.2em]
{\small Email: \texttt{f-liu@ist.osaka-u.ac.jp}}
\thanks{$^{\dagger}$Corresponding author.  }%
\thanks{*This work was supported by “Program for Leading Graduate Schools” of The University of Osaka, Japan and by JST SPRING, Japan Grant Number JPMJSP2138 and by JST BOOST, Japan Grant Number JPMJBS2402. }
}

\begin{document}

\maketitle
\thispagestyle{empty}
\pagestyle{empty}

\begin{abstract}
Assistance in collaborative manipulation is often initiated by user instructions, making high-level reasoning request-driven.
In fluent human teamwork, however, partners often infer the next helpful step from the observed outcome of an action rather than waiting for instructions.
Motivated by this, we investigate an event-driven formulation of proactive assistance, where human--object interaction outcomes initiate assistive reasoning without user-provided task specifications at inference time. 
To this end, we propose an event-driven framework that monitors workspace state changes with an event monitor and, upon event completion, extracts stabilized pre/post snapshots that characterize the resulting state transition. 
A frozen pretrained Vision-Language Model (VLM) then uses its semantic priors to infer the task context, decide whether assistance is appropriate, and, when needed, generate a sequence of assistive actions from the observed transition. 
To make outputs executable and verifiable, we restrict actions to a set of action primitives and reference objects via integer IDs.
We evaluate the same framework across three distinct real world tabletop collaboration tasks without task-specific training or fine-tuning. 
The event-driven framework achieves performance comparable to variants given user instructions. 
\end{abstract}

\section{Introduction}
\label{sec_intro}
    Recent advances in embodied intelligence have enhanced robots' ability to interpret task instructions, ground them in perception, and generate executable actions in diverse manipulation settings~\cite{kim2024openvla, black2024pi0}. 
    However, instruction following mainly addresses what to do once a specification is provided.
    To move toward fluent teamwork, a robot should be able to decide when to intervene and what assistance to provide by inferring human intention, the unobserved internal states or goal-oriented constructs that underlie observable human behavior~\cite{hoffman2024inferring}. 
    Recent reviews further suggest that proactive interaction can be facilitated by anticipation from observations and contextual information, often without explicit input from a human~\cite{den2024what}. 
    
    \begin{figure}[t!]
        \centering
        \includegraphics[width=\linewidth]{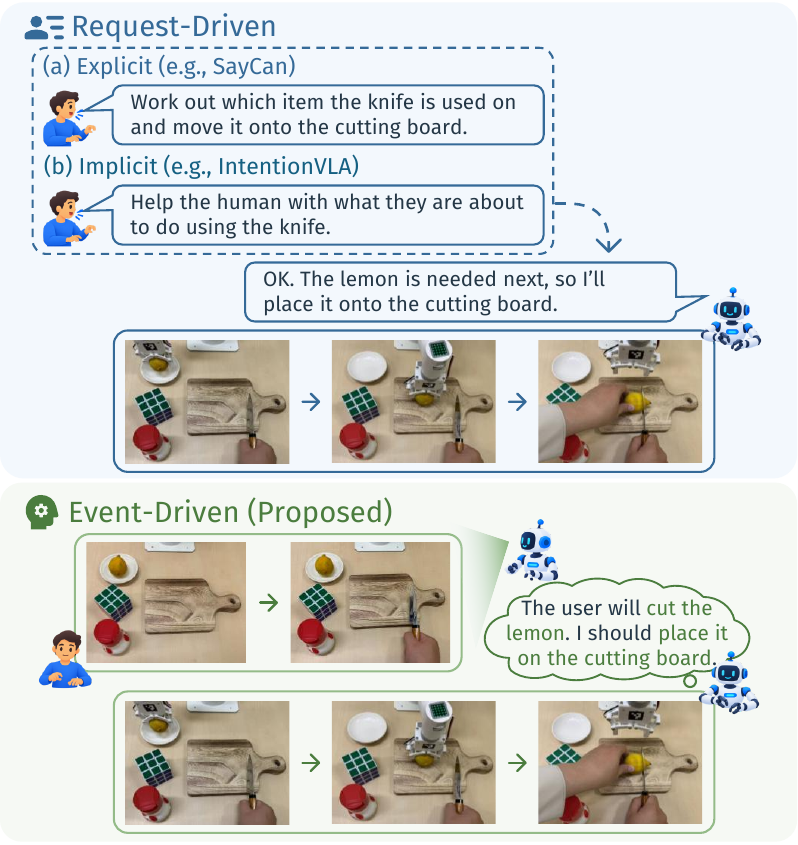}
        \caption{Motivation: request-driven vs.\ event-driven assistance. Request-driven systems initiate planning from requests provided by the user~\cite{ahn2022saycan, chen2025intention}. Our event-driven setting instead triggers on human--object state transitions to infer the user’s goal and act without an additional user request.}
        \label{fig_motivation}
    \end{figure}
    
    Translating this vision into practice reveals a bottleneck: many assistive systems remain request-driven, meaning that robot actions are initiated only after the human issues a request to the robot. 
    Requests can be explicit, for instance a natural language command that specifies what to do~\cite{mahadevan2024generative, dai2024think}, or implicit, conveyed through underspecified verbal instructions~\cite{chen2025intention} or instructive nonverbal cues~\cite{tay2026intent}.
    Even with large language models (LLMs) and vision-language models (VLMs), whose contextual reasoning can support goal inference~\cite{zhang2023large}, high-level reasoning in many recent embodied systems is still initiated from a preceding instruction or user cue~\cite{chen2025intention}. 
    Consequently, interaction often remains in a request--response loop, which limits proactive assistance when step-by-step guidance is absent. 
    
    Motivated by this gap, we move from request-driven assistance to an event-driven proactive assistance setting. 
    In human-human teamwork, partners often do not wait for explicit instructions after every step. 
    Instead, they use the observed outcome of an action to infer the likely next sub-goal and decide whether to assist. 
    Motivated by this interaction pattern, we define an event as the human--object interaction interval initiated by the human that induces a sustained workspace state change. 
    While prior work has used workspace changes for adaptive planning under a shared task description~\cite{cai2025vision}, we further use event completion itself to initiate assistance without such a specification. 
    Under this framework, the robot monitors the workspace for human--object interaction events and forms stabilized pre/post observations representing each resulting workspace state transition. 
    At event completion, it queries a VLM with the stabilized pre/post evidence to infer a task-level goal and plan an appropriate assistive response, without requiring an additional user request specifying the next step. 
    This event interface allows a frozen pretrained VLM to perform the required semantic reasoning without additional task-specific training. 
    To improve reliability and mitigate errors from generative planning, we incorporate multi-layer safeguards in the control loop, including schema-constrained outputs, ID-grounded references, and local verification. 
    We validate the framework on a real tabletop manipulation platform and compare event-driven assistance with request-driven variants under a shared evaluation protocol. 
    Our contributions are threefold: 
    \begin{itemize}
        \item \textbf{A paradigm shift from request-driven assistance to event-driven proactive assistance.}
        We cast proactive assistance in our setting as an event-driven process in which robot actions are initiated by workspace state transitions caused by human--object interactions, rather than user-provided requests or task specifications. 
    
        \item \textbf{An embodied framework for event-driven assistive reasoning and execution.} 
        We implement this formulation with runtime event detection, compact pre/post context selection, pretrained VLM reasoning, and an ID-grounded action interface with local verification.
        
        \item \textbf{A multi-task evaluation of effectiveness and adaptability.}
        We validate our framework on multiple real-world collaborative tasks, demonstrating effectiveness across different task settings and performance comparable to variants provided with user instructions.
    \end{itemize}
     
\section{Related Work}
\label{sec_related_work}

\subsection{Request-Driven Assistance}
    \label{sec_related_request_driven}
    Many collaborative robot systems remain request-driven: even as foundation models improve the perception of human behavior and contextual reasoning, robot actions are still typically initiated from explicit or inferred user requests. 
    
    \subsubsection{Explicit Request}
        Systems based on explicit requests initiate robot actions from user-provided commands or task specifications. 
        Language-to-plan pipelines condition action generation on the given instruction and ground candidate steps for execution~\cite{ahn2022saycan}, while other systems translate instructions into executable code or task structures combine visual perception~\cite{zhang2023panoptic, zhang2024panoptic} with reusable robot skills~\cite{liang2023code, mahadevan2024generative, hoffmeister2025towards}. 
        Interactive feedback and execution-time refinement can further adapt robot behavior when uncertainty or failures arise~\cite{dai2024think, hoffmeister2025towards}. 
    
        Recent embodied foundation models substantially expand instruction-conditioned manipulation through multimodal reasoning and grounded execution~\cite{zitkovich2023rt2,driess2023palme,huang2023voxposer}. 
        Recent VLA systems further extend instruction modalities and generalist manipulation capabilities~\cite{zhao2025vlas,wen2025dexvla,geminirobotics2025}.
        Despite differences in architecture and execution interface, these systems still typically initiate high-level reasoning from an instruction or stated goal. 

    \subsubsection{Implicit Request}
        Other systems operate from a user request, control cue, or task framing that indicates an intended goal without fully specifying the robot action. 
        The robot must therefore combine this information with observed context to infer appropriate assistance. 
        Prior work resolves ambiguous language using physical context, preceding human actions, or situational reasoning~\cite{wan2022handmethat,park2024clara}, while recent work using foundation models infers latent human goals or task structure from under-specified instructions to guide planning~\cite{wan2025infer, huang2026graphcotvla, chen2025intention}. 
        Nonverbal or partial control signals can serve a similar role: gaze is combined with scene context to infer manipulation intent~\cite{tay2026intent}, while shared-autonomy systems combine partial user control with environmental observations to infer or refine user goals~\cite{yow2024shared, belsare2025toward, liu2025casper}. 
        Related VLM work further demonstrates contextual inference of latent goals, human states, or manipulation intent from visual or multimodal observations~\cite{zhang2024microemo, chi2025vlmdm, wang2026roboomni, zhang2026acoustemo}. 
        Complementary work captures individual differences in human judgments through annotator-specific modeling~\cite{zhang2026qumab} and learning from incomplete annotations~\cite{zhang2026simlabel}. 
        Collaborative systems can likewise combine shared task framing with changing workspace observations to infer human intent and update robot actions~\cite{cai2025vision}. 
        Across these settings, contextual observations complement an already supplied but incomplete request, cue, or task framing, rather than determining when assistance reasoning is initiated. 

    Despite increasingly capable intent inference, request-driven systems still depend on a user request, control cue, or task framing to initiate assistance. 
    This leaves the initiative primarily on the human side, even when the robot can infer substantial task context on its own. 
    To support proactive collaboration closer to human--human teamwork, robots should also be able to initiate assistance from the observed outcomes of human interaction without waiting for an additional request. 
    This motivates the event-driven assistance setting considered here. 
    
\subsection{Event-Driven Assistance}
    \label{sec_related_event_driven}
    Prior work has taken steps toward proactive collaboration by using observed human activity and context to derive cues for assistance without relying on a user request~\cite{patel2023predicting}. 
    Such systems typically model semantic aspects of the ongoing interaction, including anticipated human--object interactions~\cite{mascaro2023hoi4abot}, action completion~\cite{lazzari2025pace}, human intent or latent goals~\cite{kedia2024interact, puig2023nopa}, or broader mental states~\cite{zhang2025autotom}.  
    These estimates are then used to determine or adapt robot assistance.

\begin{figure*}[t]
    \centering
    \includegraphics[width=\linewidth]{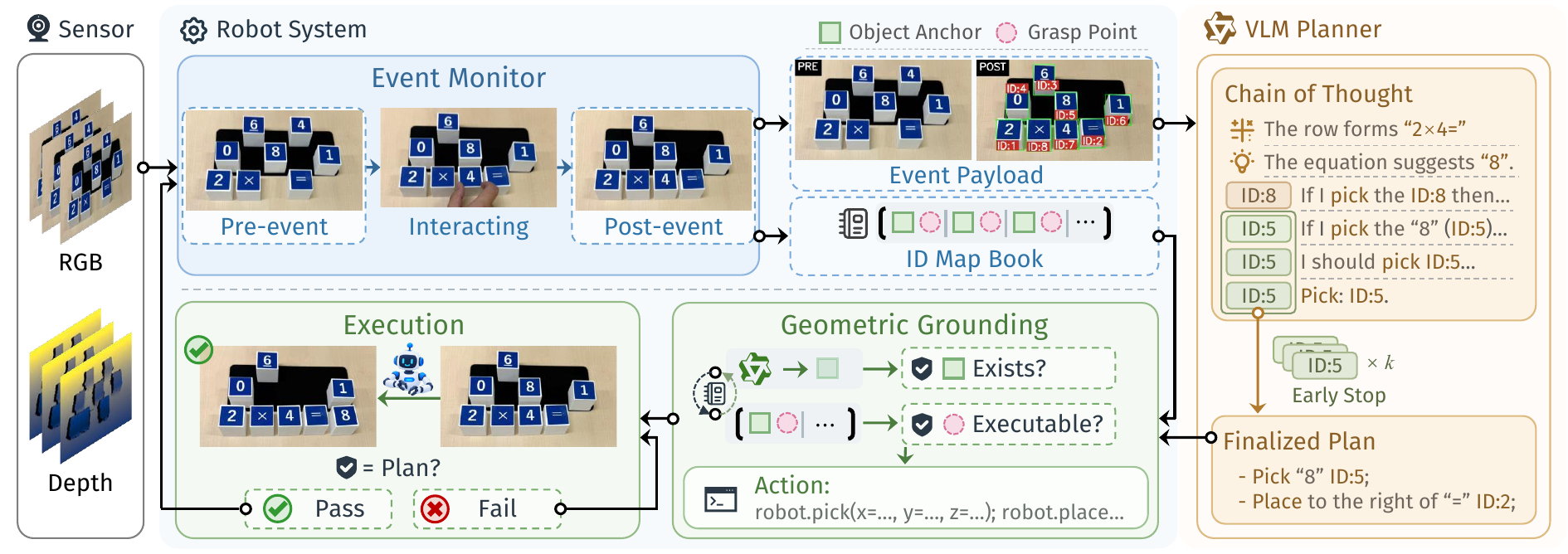} 
    \caption{System overview. Human-induced workspace transitions trigger event processing that produces pre/post evidence and an ID map book. Without an explicit user request, the VLM reasons over this event payload with early stopping to finalize a symbolic plan, which is geometrically grounded, executed, and verified before updating the workspace baseline.} 
    \label{fig_overview}
\end{figure*}

    Across these approaches, proactive assistance is mediated by semantic representations of human behavior, such as human--object interactions, action progress, intent, goals, or mental states.  
    Many methods use dedicated predictors or structured inference models to obtain these representations. 
    We explore a complementary route in which physical interaction outcomes provide a common event interface that a frozen pretrained VLM can interpret without additional task-specific training. 
    Related work has used pre/post observations to recognize manipulation outcomes or responded to workspace changes by adapting existing plans~\cite{goko2024task, li2021reactive}. 
    In our setting, a completed human--object transition serves both as the event that initiates assistance and as the paired evidence from which the VLM infers the next assistive action. 
\section{Methodology}
\label{sec_method}
\subsection{Problem Setting: Event-Driven Proactive Assistance}
    \label{sec_method_formulation}
    We focus on proactive assistive manipulation in human--robot shared workspaces.
    An interaction event is a human--object interaction interval that induces a sustained workspace state change, with onset and offset times detectable from changes in the observed workspace state. 
    We assume that the outcome of such interactions provides task-relevant evidence. 
    Accordingly, we define the assistance objective at the task-level from the observed pre/post state transition, such as completing an object placement or establishing a spatial relation. 
    Once an event terminates, the system infers an appropriate assistive response and decides whether to act or wait. 
    Triggering at event completion provides more informative evidence than intermediate frames and supplies sufficient context for inferring an appropriate assistive response. 

\subsection{System Architecture}
    \label{sec_method_architecture}

    Completed human--object interactions serve two roles in our framework: event completion determines when assistance reasoning is invoked, while the paired pre/post observations expose the semantic consequence of the interaction. 
    This context allows a pretrained VLM, used without task-specific fine-tuning, to infer an appropriate assistive response from the observed state transition. 
    To connect this event-level reasoning to robot control, the planner operates over a structured event payload and returns ID-indexed actions from a restricted primitive set, while the robot system handles metric grounding and execution checks. 
    
    Fig.~\ref{fig_overview} illustrates the architecture, consisting of a robot system and an event-triggered VLM planner. 
    The robot system monitors the RGB--D stream, maintains the event state, and forms a pre/post snapshot pair $(I^{-}, I^{+})$ when an interaction event completes. 
    The post-event scene is parsed into an ID-indexed object map $\mathcal{T}^{+}$, with the IDs overlaid on $I^{+}$ for VLM reference. 
    Given $(I^{-}, I^{+})$, the VLM planner returns an ID-indexed symbolic plan over the allowed primitives. 
    The robot controller then validates and grounds the plan using $\mathcal{T}^{+}$, executes it, and verifies the resulting workspace state. 
    We next detail the three stages of this pipeline: event processing and representation, event-driven VLM planning, and geometric grounding and execution. 
    
\subsection{Event Processing and Representation}
    \label{sec_method_local_pre}
    \subsubsection{Event Monitoring}
        \label{sec_method_local_event}

        Event progress is tracked through three states: pre-event $\rightarrow$ interacting $\rightarrow$ post-event. 
        In the pre-event state, the system maintains a stable workspace height map $H^{-}$ and its RGB observation $I^{-}$, with height maps expressed relative to the table plane. 
        For each incoming height map, change statistics are computed over the union of the current and reference regions within a fixed height band above the table, capturing both removed and newly occupied areas while excluding operator-side intrusions. 
        The system computes an instantaneous change ratio $e_t$ between consecutive observations and a cumulative change ratio $d_t$ relative to $H^{-}$, with the height threshold for $e_t$ adapted to stationary noise. 
        The system enters interacting when either $e_t>\theta_e$ or $d_t>\theta_d$ persists for the required evaluations, and enters post-event when $e_t<\theta_q$ for a fixed dwell. 
        Renewed disturbance returns it to interacting. 
        In the post-event state, the changed region relative to $H^{-}$ is evaluated across consecutive observations for spatial overlap, geometric height consistency, intrusion overlap, and valid depth. 
        Satisfying these conditions confirms event completion, while an empty region or timeout returns the system to pre-event without invoking the VLM planner. 
        The confirmed stable observation becomes $I^{+}$, so $(I^{-},I^{+})$ spans two stable workspace states without intermediate manipulation frames. 
        Monitoring then pauses and resumes using the resulting stable workspace state as the new baseline. 

    \subsubsection{Post-Event Parsing and Object Map Construction}
        \label{sec_method_local_map}
        After an event terminates, the robot system parses the post-event snapshot $I^{+}$ to obtain object instances for manipulation and grounding. 
        For efficient instance labeling, we use FastSAM, followed by RGB--D refinement relative to the table plane. 
        The resulting instances are represented as
        \begin{equation}
            \mathcal{D}(I^{+}) = \{(b_k,m_k)\}_{k=1}^{K},
        \end{equation}
        where $b_k$ and $m_k$ denote the bounding box and instance mask of object $k$. 
        Each instance is assigned an integer ID and associated with its geometric information, yielding an ID-indexed object map
        \begin{equation}
            \mathcal{T}^{+}[\textit{id}] = (b,m,\mathbf{p},\mathcal{G}),
        \end{equation}
        where $\mathbf{p}$ is the center of the estimated top surface and $\mathcal{G}$ contains its extent and principal directions.
        The map $\mathcal{T}^{+}$ is retained by the robot system for resolving IDs returned by the planner and geometric grounding.
        For VLM planning, the snapshots are marked as PRE and POST, with object IDs overlaid only on $I^{+}$ so that they refer unambiguously to the executable scene.
        The IDs are rendered within their corresponding bounding boxes while avoiding overlaps between labels.
        The robot system then forms the event payload $\mathcal{E}=(I^{-},I^{+})$ and provides it to the VLM planner. 

\subsection{Event-Driven VLM Planning}
    \label{sec_method_vlm}
    Given the event payload $\mathcal{E}$, the VLM interprets the state transition to infer an appropriate assistive goal and grounds it to the identified objects. 
    A fixed system prompt specifies the available action primitives and structured output format. 
    The resulting symbolic plan uses only object IDs and contains \texttt{pick}, \texttt{place}, or \texttt{wait}, with placement defined by a reference ID and qualitative spatial relation. 

    To avoid overthinking after a target has stabilized, the generated chain-of-thought (CoT) is continuously monitored for manipulation-related statements. 
    Inspired by prior work on CoT convergence~\cite{mao2026early}, a parser extracts the target object ID from sentences containing verbs such as \emph{move}, \emph{place}, or \emph{pick}, and tracks repeated IDs across the reasoning trace. 
    Questions, speculative phrasing (e.g., \emph{maybe}, \emph{perhaps}, or \emph{what if}), and meta-level references are ignored, while pronouns inherit the most recent candidate ID. 
    Let $c_1,c_2,\ldots$ denote the resulting sequence of candidate IDs. 
    Generation is stopped when the same ID appears for $k=3$ consecutive candidates, $c_{n-k+1}=\cdots=c_n$. 
    The converged ID is then fixed in a separate finalization call that reuses the generated reasoning as context and completes the remaining plan fields without revisiting the selected target. Consequently, \texttt{wait} is no longer available after early stopping.  
    If reasoning terminates normally without producing any candidate ID, the model's original structured outcome is retained. 
    When only one candidate ID is observed and the final plan selects it, the plan is accepted without further interaction. 
    
    To avoid premature target decisions, other normally terminated but non-converged candidate sequences are treated as unresolved target decisions and invoke a brief QA step. 
    The QA menu contains the distinct candidate IDs extracted from the CoT together with an additional \texttt{wait} option. 
    If the human selects an ID, that ID is fixed through the same finalization call, again reusing the preceding reasoning; selecting \texttt{wait} directly yields a wait outcome. 
    The selected response is then used to produce the final structured outcome consumed by the grounding and execution stages. 

\begin{figure*}[!t]
    \centering
    \includegraphics[width=\linewidth]{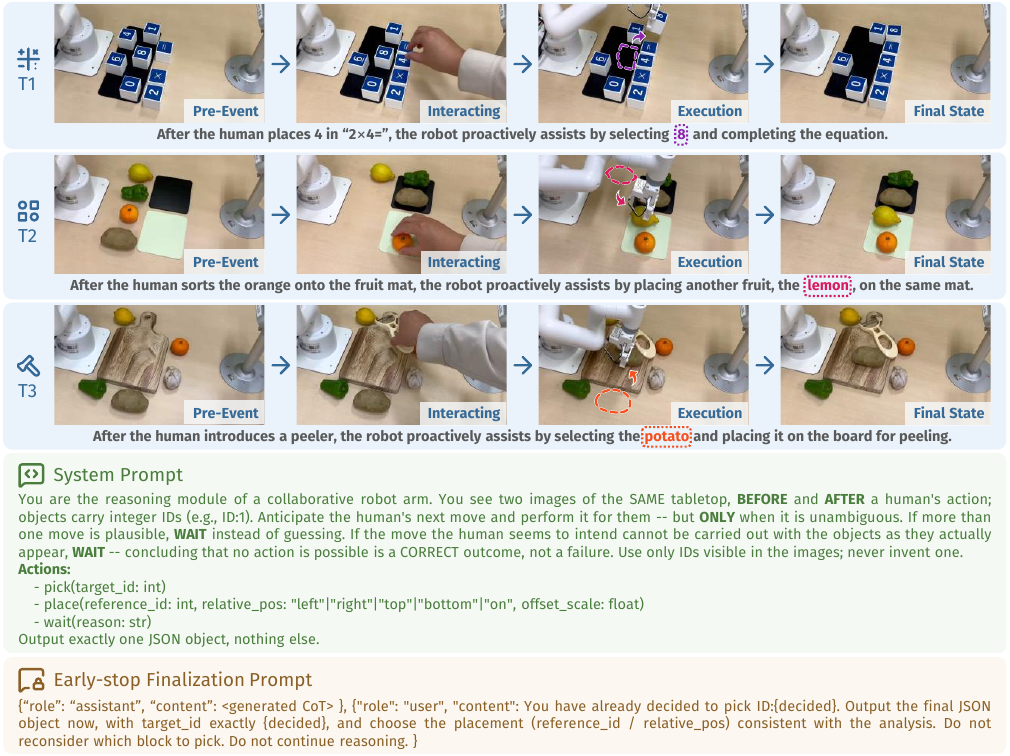}
    \caption{Experimental scenarios and VLM prompts. Top: representative interaction sequences for the three tasks. Middle: the common system prompt. Bottom: the early-stop finalization prompt, which retains the generated reasoning while fixing the selected target ID. The JSON output schema is omitted for brevity.}
    \label{fig_exp_overview}
\end{figure*}

\subsection{Geometric Grounding and Execution}
    After the VLM planner returns an action list, the robot controller validates the primitive type and referenced IDs, then binds each action to the corresponding geometry in $\mathcal{T}^{+}$. 
    The resulting metric targets are transformed into the robot base frame using the calibrated camera-to-base transform. 
    
    For \texttt{pick(target\_id)}, the controller resolves the target ID and uses the planar coordinates of its top-surface center $\mathbf{p}$ as the grasp location.
    The grasp height is determined by the controller configuration, with a fixed downward gripper orientation. 
    The resulting grasp target is executed through MoveIt. 
    For \texttt{place(reference\_id, relative\_pos, offset\_scale)}, the controller resolves the reference center $\mathbf{p}_r$ and converts the qualitative relation into a metric placement target. 
    For a directional relation $q$,
    \begin{equation}
        \mathbf{p}_{\mathrm{place}} = \mathbf{p}_r + \left(s_r\alpha + \delta\right)\mathbf{u}_q ,
    \end{equation}
    where $\alpha$ denotes \texttt{offset\_scale}, $s_r$ is the full extent of the reference object's top surface along direction $\mathbf{u}_q$, and $\delta$ is a local safety margin. 
    Directional relations, such as \texttt{left} and \texttt{right}, are interpreted in the camera view, while \texttt{on} is handled separately by centering the target over the reference object and adjusting the release height. 
    Before execution, the action sequence is validated, and any \texttt{place} without a preceding \texttt{pick} is rejected. 
    The resulting placement target is executed through the same MoveIt motion planner. 
    Thus, the VLM plan remains symbolic while metric grounding and motion execution are handled by the robot system. 
    
    After execution, the system captures a refreshed observation and verifies whether the qualitative relations specified by the symbolic plan hold in the resulting scene. 
    A failure is declared when expected relations are violated or referenced objects cannot be recovered. 
    If verification fails, the plan may be re-executed once and verified again. 
    The settled workspace then provides the baseline for subsequent event monitoring. 
    
\section{Experiments}
\label{sec_evaluation}
In this section, we evaluate the effectiveness of our event-driven proactive assistance framework on three tabletop collaboration tasks.
Video demonstration is provided in the \textbf{supplementary material}.

\subsection{Experimental Setting}
\label{sec_exp_setting}
    \subsubsection{Task Design}
        We evaluate three tabletop collaboration tasks (Fig.~\ref{fig_exp_overview}): equation completion (T1), tabletop sorting (T2), and tool introduction (T3). 
        Each event uses a distinct object layout, while T1 varies the candidate block composition. 
        In T1 (60 events), the human adds one of five candidate blocks to a partial single digit arithmetic expression covering addition, subtraction, multiplication, and division, and the robot completes it. 
        In half of the scenes, the same digit also appears in the expression while the correct block to move remains unique. 
        In T2 (30 events), the candidate items are a lemon, an orange, a potato, and a green pepper. 
        The human places three to establish the fruit/vegetable mat mapping, and the robot places the remaining item. 
        In T3 (30 events), the human introduces a peeler among several candidate ingredients, and the robot moves the uniquely applicable potato onto the cutting board. 
        Across 120 events, outcomes are classified as autonomous success, valid QA, or failure; appropriate denotes autonomous success plus valid QA. 
    
    \subsubsection{Implementation}
        Experiments use an Elephant Robotics MyCobot 280 Arduino with a RealSense D435i. 
        Perception, event processing, and robot control run on an NVIDIA Jetson Orin Nano 8GB, while VLM inference runs on four NVIDIA A6000 GPUs using Qwen3.5-122B-A10B-FP8 served with vLLM. 
        VLM requests and responses are exchanged via ZeroMQ (ZMQ). 
        Decoding uses temperature 0.1, top-$p$ 0.95, and up to 8192 output tokens; the finalization call suppresses further reasoning and is limited to 1200 output tokens. 
        Multi-token prediction is enabled with two speculative tokens for faster inference. 
        Apart from the intended experimental differences, all variants use the same model and action interface. 
        For event monitoring, $\theta_e=\theta_d=0.010$, with persistence requirements of 2 and 3 successive evaluations for $e_t$ and $d_t$, respectively, while $\theta_q=0.008$ with a 0.7\,s dwell. 
        The monitored height band is 6--80\,mm above the table. 
        For $e_t$, the height threshold is adapted from the stationary 99.7th percentile and clipped to 2--12\,mm; $d_t$ and changed-region extraction use a fixed 10\,mm threshold. 
        Event completion requires IoU $\geq0.60$ for 3 checks, intrusion overlap $\leq0.15$, valid depth $\geq0.55$, and height variation within twice the stationary standard deviation. 
        Across all experiments, we observed neither spurious planner invocations nor event fragmentation. 

\subsection{Baselines}
    \label{sec_exp_variants}
    The request-driven variants add only a user instruction to the corresponding system prompt, adapting prior task-specification settings to a common planner and observation interface for controlled comparison. 
    \emph{Explicit Request}~\cite{liang2023code, huang2023voxposer} instead specifies the task rule: ``Compute the equation and move the resulting number to the end of the equation.'' for T1, ``Move one of the remaining items onto the mat where an item of the same category has already been placed.'' for T2, and ``Work out which item on the table the tool the human just put down is used on, and if there is such an item, move it onto the cutting board.'' for T3. 
    \emph{Implicit Request}~\cite{wan2025infer, cai2025vision} uses broad collaborative goals: ``Work together with the human to complete this equation.'' for T1, ``Help the human organize the items on the table.'' for T2, and ``Help the human with what they are about to do.'' for T3. 
    Both use the same pre- and post-event observations as ours and are also evaluated without pre-event context.

\subsection{Experimental Results}
\label{sec_exp_results}
    \subsubsection{Qualitative Results}
        Fig.~\ref{fig_exp_overview} shows representative interaction sequences across the three tasks. 
        The robot completes an arithmetic expression in T1, places the object according to the inferred category mapping in T2, and identifies an object through its functional relation to the introduced tool in T3.  
        Although these tasks require different forms of relational reasoning, the framework consistently turns the observed workspace transition into proactive assistive execution. 

        \begin{table*}[t]
            \centering
            \setlength{\tabcolsep}{2pt}
            \caption{Benchmark results across the three tasks. \emph{Appropriate} (\%) is the sum of autonomous success and valid QA; \emph{Succ.\,/\,QA\,/\,Fail} (\%) reports autonomous success, valid QA, and failure as percentages; \emph{latency} is reported in seconds. }
            \label{tab_benchmark}
            \begin{tabular}{l ccc ccc ccc}
                \toprule
                \multirow{2}{*}{Method} & \multicolumn{3}{c}{T1: Equation completion} & \multicolumn{3}{c}{T2: Tabletop sorting} & \multicolumn{3}{c}{T3: Tool introduction} \\
                \cmidrule(lr){2-4} \cmidrule(lr){5-7} \cmidrule(lr){8-10}
                & Appropriate & Succ.\,/\,QA\,/\,Fail & Latency & Appropriate & Succ.\,/\,QA\,/\,Fail & Latency & Appropriate & Succ.\,/\,QA\,/\,Fail & Latency \\
                \midrule
                \textbf{Ours} & \textbf{100.0} & \textbf{95.0}\,/\,5.0\,/\,\textbf{0.0} & 8.6 & 80.0 & \textbf{80.0}\,/\,0.0\,/\,20.0 & 8.6 & \textbf{100.0} & \textbf{80.0}\,/\,20.0\,/\,\textbf{0.0} & 11.1 \\
                \quad w/o Pre-event Context & 93.3 & 91.7\,/\,1.7\,/\,6.7 & 6.8 & 60.0 & 53.3\,/\,6.7\,/\,40.0 & 7.4 & 90.0 & \textbf{80.0}\,/\,10.0\,/\,10.0 & 9.4 \\
                \midrule
                Explicit Request~\cite{liang2023code, huang2023voxposer} & 98.3 & 91.7\,/\,6.7\,/\,1.7 & 7.2 & \textbf{86.7} & \textbf{80.0}\,/\,6.7\,/\,\textbf{13.3} & 7.2 & \textbf{100.0} & 53.3\,/\,46.7\,/\,\textbf{0.0} & 5.0 \\
                \quad w/o Pre-event Context & 90.0 & 81.7\,/\,8.3\,/\,10.0 & \textbf{4.5} & 70.0 & 60.0\,/\,10.0\,/\,30.0 & 6.8 & \textbf{100.0} & 76.7\,/\,23.3\,/\,\textbf{0.0} & \textbf{4.8} \\
                \midrule
                Implicit Request~\cite{cai2025vision, wan2025infer} & 98.3 & \textbf{95.0}\,/\,3.3\,/\,1.7 & 8.6 & 76.7 & 76.7\,/\,0.0\,/\,23.3 & 12.8 & 90.0 & 63.3\,/\,26.7\,/\,10.0 & 14.6 \\
                \quad w/o Pre-event Context & 90.0 & 90.0\,/\,0.0\,/\,10.0 & 5.0 & 83.3 &76.7\,/\,6.7\,/\,16.7 & \textbf{5.6} & 90.0 & \textbf{80.0}\,/\,10.0\,/\,10.0 & 7.4 \\
                \bottomrule
            \end{tabular}
        \end{table*}

        \begin{table}[t]
            \centering
            \setlength{\tabcolsep}{6pt}
            \caption{Ablation study on the proposed components. All values are reported as \emph{Appropriate} (\%).}
            \label{tab_ablation}
            \begin{tabular}{l ccc}
                \toprule
                Variant & T1 & T2 & T3 \\
                \midrule
                \textbf{Ours (Full)} & \textbf{100.0} & \textbf{80.0} & \textbf{100.0} \\
                w/o Event Triggering ($\tau=2$\,s) & 23.3 & 3.3 & 60.0 \\
                w/o Event Triggering ($\tau=1$\,s) & 20.0 & 0.0 & 86.7 \\
                w/o Early Stopping & 86.7 & 60.0 & 70.0 \\
                \bottomrule
            \end{tabular}
        \end{table}

    \subsubsection{Quantitative Results}
        \begin{figure}[t]
            \centering
            \includegraphics[width=\linewidth]{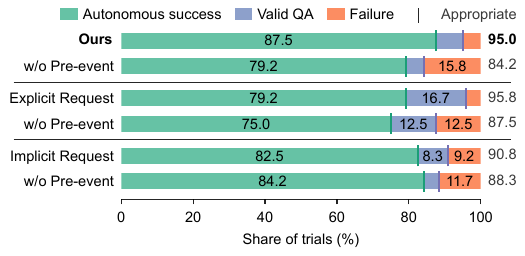}
            \caption{Overall outcomes across the evaluated events. \emph{Appropriate} is the sum of autonomous success and valid QA; a QA is valid only when the correct option is retained and leads to a correct plan.}
            \label{fig_outcome}
        \end{figure}

        Ours achieves 95.0\% appropriate over the 120 evaluated events, as summarized in Fig.~\ref{fig_outcome}. 
        Explicit request and implicit request achieve 95.8\% and 90.8\%, respectively, with no significant difference in failure rate from ours in paired exact McNemar tests (5.0\% vs.\ 4.2\%, $p=1.0$; 5.0\% vs.\ 9.2\%, $p=0.27$). 
        Thus, when a correct assistive action is available, inferring it from the observed event without a user request achieves comparable appropriate rates to the request-driven variants in the evaluated tasks. 
        
        The results further show that additional request information does not consistently improve assistance. 
        In T1, all methods achieve near-saturated appropriate rates, leaving little room for additional target information to help. 
        In T2, even with the task rule provided, the VLM can still fail to infer the correct relative placement, for example by selecting the correct object but placing it outside the target mat. 
        Thus, task instructions do not eliminate errors in relative placement reasoning. 
        In T3, explicit request achieves a similar appropriate rate to ours but enters QA more often. 
        Its reasoning trajectories are substantially shorter, and every QA case includes a candidate switch; the shorter generation may leave insufficient room to rebuild a stable run satisfying the early-stopping threshold. 
        In contrast, our design provides event-aligned pre/post context from which the required assistive action can be inferred. 
        
        During the experiments, we observed that request-driven performance was sensitive to instruction wording. 
        Before fixing the prompts used in the main comparison, we tested semantically similar wording variants to remove linguistic ambiguity. 
        In T2, the explicit request phrase ``same kind of item'' was often interpreted as requiring an identical object and caused unnecessary abstention; replacing it with ``same category'' reduced such errors from 28\,/\,60 to 6\,/\,60. 
        Table~\ref{tab_benchmark} reports the revised wording. 
        A separate T1 wording test showed a similar effect: a target description such as ``the block showing 6'' could also refer to the identical digit already placed in the equation, while adding ``in the back row'' eliminated these errors (20\,/\,60 $\rightarrow$ 0\,/\,60). 
        Thus, request-driven performance can depend not only on the information provided but on how it is expressed, whereas ours introduces no such user instruction variable. 
    
    \subsubsection{Ablation Study}
        We use controlled variants to examine event triggering, pre/post-event visual context, and generation termination. 
        \emph{w/o Event Triggering} queries the planner at fixed intervals of $\tau=2$ or $1$\,s, while \emph{w/o Pre-event Context} retains only the post-event image.
        Their system prompts differ only in the visual-input description. 
        \emph{w/o Early Stopping} disables early stopping with otherwise identical input and prompt, producing no QA outcomes. 
        Ablation results are reported in Table~\ref{tab_ablation}, with the pre-event comparison in Table~\ref{tab_benchmark}.  
    
\begin{figure}
            \centering
            \includegraphics{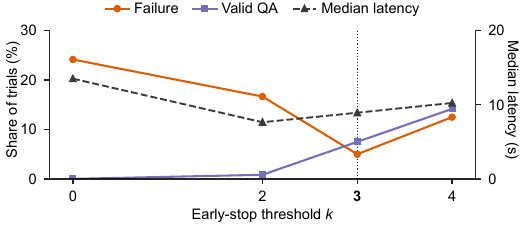}
            \caption{Sensitivity analysis of the early-stop threshold $k$. $k=3$ gives the lowest failure rate (5.0\%) with 7.5\% valid QA and a median latency of 8.93\,s. } 
            \label{fig_early_k}
        \end{figure}

        All three components contribute to performance. 
        Removing event triggering reduces appropriate from 95.0\% to 27.5\% at $\tau=2$\,s and 31.7\% at $\tau=1$\,s. 
        Without event triggering, the two observations may capture transient states while the human action is still unfolding, making the resulting plan stale by execution time. 
        Removing pre-event context lowers appropriate to 84.2\% and increases failures from 6 to 19 ($p=0.010$); abstention despite a uniquely required action increases from 2 to 14, indicating that the post-event state alone often underspecifies the observed change. 
        Removing early stopping lowers appropriate to 75.8\% and increases latency in all three tasks, indicating that stable-candidate termination improves reliability and avoids unnecessarily long generation. 
        Further testing of $k=0,2,3,4$ shows that $k=3$ gives the lowest failure rate (5.0\%) with a median latency of 8.93,s (Fig.~\ref{fig_early_k}), supporting its use as the default. 
        A consistent trend is also observed with Qwen3.8-27B. 
    
\subsection{Discussion}
    \label{sec_boundary}
    \subsubsection{Boundary Events}
        \begin{table}[t]
            \centering
            \setlength{\tabcolsep}{2pt}
            \caption{Boundary event results. All values are reported as  \emph{Appropriate} (\%).}
            \label{tab_boundary}
            \begin{tabular}{l cc}
                \toprule
                Method & Optional Intervention &\quad Required Abstention \\
                \midrule
                Ours             & 82.2 & \textbf{76.7} \\
                Explicit Request & \textbf{83.3} & 50.0 \\
                Implicit Request & 80.0 & 60.0 \\
                \bottomrule
            \end{tabular}
        \end{table}

        The main benchmark considers cases in which the scene supports a uniquely appropriate assistive action. 
        We next evaluate 150 boundary events across five 30-trial designs using the same object sets, grouped by whether action is optional or inappropriate. 
        The 90 \emph{optional intervention} cases include T2 with two or one demonstrated placements and T3 with a knife applicable to multiple targets. 
        Under the system prompt in Fig.~\ref{fig_exp_overview}, unresolved multiple candidates may return wait. 
        Thus, either a valid action or wait is accepted. 
        The remaining 60 \emph{required abstention} cases include T1 with no block that completes the equation and T3 with a scoop applicable to none of the candidate objects, and only wait is a correct autonomous response. 

        As shown in Table~\ref{tab_boundary}, the three methods achieve similar appropriate rates when intervention is optional. 
        At the action level, ours executes a pick in 45.6\% of these cases, compared with 61.1\% for explicit request and 73.3\% for implicit request, showing that the request-driven variants act autonomously more often under ambiguous evidence. 
        When abstention is required, any pick is unsupported and therefore incorrect. 
        The corresponding pick rates are 23.3\%, 50.0\%, and 40.0\%, respectively ($p=0.0052$ for explicit and $p=0.053$ for implicit versus ours). 
        For example, the planner may reuse a block already placed in the T1 equation or, in T3, select an object near the introduced tool even though the tool applies to none of the candidates. 
        Overall, these boundary cases suggest that request information increases the tendency to commit to an action under uncertain evidence, while ours remains more selective without reducing appropriateness. 

    \subsubsection{Failure Cases}
        The boundary cases above expose a remaining limitation of the VLM planner: when scene evidence is insufficient, it may still construct and commit to an executable action rather than retain wait.
        A second limitation lies in spatial reasoning.
        In T2, the VLM may identify the correct object but still infer an incorrect relative placement, occasionally placing it outside the target mat or on top of an existing object. 
        Future work will explore replacing direct VLM planning with frozen VLM representations as semantic interfaces for learned world models that verify action applicability, together with structured geometric prediction for more reliable spatial grounding. 

\section{Conclusion}
\label{sec_conclusion}
    We present an event-driven framework for proactive assistance, where robot actions are triggered by human-induced workspace state transitions rather than task instructions. 
    Across three collaborative tasks, the framework provides overall performance comparable to request-driven variants, supporting event-driven assistance as a viable alternative to request-driven interaction. 


\bibliographystyle{IEEEtran}
\bibliography{references}

\end{document}